\documentclass[10pt,twocolumn,letterpaper]{article}

\usepackage{iccv}
\usepackage{times}
\usepackage{epsfig}
\usepackage{graphicx}
\usepackage{amsmath}
\usepackage{amssymb}

\newcommand{\degree}[1]{{$#1 ^{\circ}$}}

\usepackage[breaklinks=true,bookmarks=false]{hyperref}

\iccvfinalcopy 

\begin{document}

\title{Illuminant Chromaticity Estimation from Interreflections}

\author{Eytan Lifshitz\\ 
The Hebrew University of Jerusalem\\
{\tt\small eytanl@cs.huji.ac.il}
\and
Dani Lischinski\\
The Hebrew University of Jerusalem\\
{\tt\small danix@cs.huji.ac.il}
}

\maketitle

\begin{abstract}

Reliable estimation of illuminant chromaticity is crucial for simulating color constancy and for white balancing digital images. However,
estimating illuminant chromaticity from a single image is an ill-posed task, in general, and existing solutions typically employ a variety of assumptions and heuristics.
In this paper, we present a new, physically-based, approach for estimating illuminant chromaticity from interreflections of light between diffuse surfaces. Our approach assumes that all of the direct illumination in the scene has the same chromaticity, and that at least two areas where interreflections between Lambertian surfaces occur may be detected in the image. No further assumptions or restrictions on the illuminant chromaticty or the shading in the scene are necessary. Our approach is based on representing interreflections as lines in a special 2D color space, and the chromaticity of the illuminant is estimated from the approximate intersection between two or more such lines.

Experimental results are reported on a dataset of illumination and surface reflectance spectra, as well as on real images we captured. The results indicate that our approach can yield state-of-the-art results when the interreflections are significant enough to be captured by the camera.

\end{abstract}
\section{Introduction}

Illuminant chromaticity estimation is important for simulating the color constancy property of the human visual system and for white balancing digital images. However, recovering the chromaticity of an illuminant from a single image is an ill-posed problem: since the reflectances of the scene's surfaces are not known, there is an inherent ambiguity that admits infinitely many solutions. 

In practice, single image color constancy and white balance approaches typically rely on certain assumptions and heuristics in order to estimate the illuminant chromaticity, such as restricted gamuts, the gray world or the white patch assumption, etc.~\cite{DBLP:journals/tip/GijsenijGW11}. Methods have also been proposed that leverage the availability of multiple images, \eg, \cite{DBLP:conf/iccv/PrinetLW13} and/or specular highlights \cite{DBLP:conf/cvpr/TanNI03,yoon2005illuminant}.


In this paper, we propose a new, physically-based, approach for estimating illuminant chromaticity from diffuse interreflections of light between pairs of nearby surfaces in the scene. This approach works on a single image of a scene assuming that all of the direct illumination in the scene has the same chromaticity, and that two or more diffuse interreflections are present and can be detected by the camera's sensor. No further arbitrary or unrealistic assumptions about the illuminant or the shading in the scene are made. Our approach is applicable to scenes where no highlights are visible, or where highlights are visible but saturated. Diffuse interreflections abound in real scenes, however their small relative magnitude makes them difficult to capture. Nevertheless, with the emergence of new sensitive cameras, such as the Modulo camera \cite{DBLP:conf/iccp/ZhaoSFYR15} with long exposure, we expect that it would become feasible to capture interreflections in just about any scene.

In an early work, Funt \etal~\cite{DBLP:journals/ijcv/FuntDH91} demonstrated the usefulness of interreflections for color constancy. They use finite-dimensional linear models to model the illumination and the surface reflectance spectra. The weights of these linear models may be recovered from a sufficient number of measurements in the vicinity of interreflections by solving a non-linear system of equations. In contrast, our approach does not make these assumptions, and recovers the chromaticity of the illuminant using a simple geometric computation in a 2D color space. The approach is shown to yield good estimation accuracy on a test set of real images.

We first show that from a measurement of a pure interreflection, we are able to resolve the reflectance-illuminant ambiguity and correctly recover the color of the illuminant.
However, in real images, the light leaving a surface is a combination of the reflection of direct and indirect illumination, and the relative magnitudes of these two components are not known. This presents us with another ambiguity. However, in areas where the indirect illumination is primarily due to an interreflection, we are able to express all the possible combinations of direct and indirect illumination colors as a line in a special 2D chromaticity plane. Thus, the illuminant chromaticity may be estimated by obtaining and intersecting two or more such lines.



Our approach assumes that all of the direct illumination in the scene has the same chromaticity. We further assume that it is possible to identify at least two surface pairs in the scene, such that on both surfaces of each pair one can measure the reflection of direct illumination only, and on one surface of each pair one can also measure the color resulting from the combined effect of direct and indirect illumination. We assume that on these surfaces the effect of indirect illumination arriving after more than one bounce of light is negligible.
The automatic detection of the regions in which these measurements are to be taken is outside the scope of the current work. 

The accuracy of the proposed approach is evaluated through computer simulations that use a large dataset of measured illuminant and reflectance spectra, as well as a series of photographs of a real scene, where the ground truth illuminant chromaticity is obtained using a gray card.

\section{Related Work}

Computational color constancy is an important task in computer vision and image processing, which has attracted a considerable amount of research attention. A comprehensive survey can be found at \cite{DBLP:journals/tip/GijsenijGW11}. Existing methods can be roughly divided into 3 groups: low-level statistics-based methods, learning-based methods and physically-based methods.

Low-level statistics-based methods rely on heuristics such as the Gray-World assumption \cite{buchsbaum1980spatial}, as well as many other more sophisticated heuristics; such heuristics are based on assumptions which do not always hold.

Learning-based methods leverage the availability of data and harness machine learning algorithms to solve the problem. They require many images for training, but they achieve state-of-the-art results, such as in \cite{DBLP:journals/corr/BiancoCS15}. Recently, Shi \etal \cite{DBLP:conf/eccv/ShiLT16} presented a neural network which consists of two sub-networks. The first network, HypNet (Hypotheses Network), has two branches. This network learns to give two estimates of the illuminant chromaticity. Each branch is specialized in different settings, so they complement each other. The second network, SelNet (Selection Network), learns to decide which of the two hypotheses from HypNet is more likely to be accurate. The authors report a significant accuracy improvement, compared to previous work.

The physically-based approach presented by Shafer \cite{DBLP:journals/ijcv/KlinkerSK90}
tries to find some physically-based hints for the illuminant chromaticity, while avoiding making overly restrictive assumptions. This approach includes the use of specular highlights as did Tan \etal \cite{DBLP:conf/cvpr/TanNI03} and Yoon \etal \cite{yoon2005illuminant}. Both of the above methods use physical models to derive straight lines in some color space, which go through the illuminant chromaticity. These works provided inspiration for our work, which also makes use of color lines. The main shortcoming of these methods is that they are not always applicable: specular highlights are not always present in the image, and when they do appear, they are often clipped. 

Another physically-based approach was introduced by Finlayson \etal \cite{finlayson2001solving}. They assume the model of black-body radiators to exploit the fact that all illuminant chromaticities are located on the Planckian locus. Our method does not assume anything about the illuminant chromaticity, so it can work on illuminants which aren't black-body radiators. There are also methods, such as \cite{DBLP:conf/iccv/PrinetLW13}, which use multiple images, and can work even in a scene with multiple illuminants. Our approach assumes a single illuminant chromaticity, but it is applicable to a single image.

Previous work on interreflection was done by Nayar \etal \cite{nayar1992colored, DBLP:conf/iccv/NayarIK90} who used interreflections in order to recover shape. Funt \etal \cite{DBLP:journals/ijcv/FuntDH91} use mutual interreflection to estimate the illuminant spectrum rather than just its chromaticity as we do. This method is based on the finite-dimensional linear model presented by Maloney \cite{maloney1986evaluation}, who pointed out the fact that illumination and surface reflectance spectra can be approximated with low dimensional linear spaces. But they are limited by the approximation error of this approach. Moreover, they assume that the illumination and the surface reflectance spectra are a linear combination of only 3 basis functions. But recently Wug Oh \etal \cite{wug2016yourself} claimed that surface reflectances are 8-dimensional and illumination spectra are 4-dimensional for outdoors and 6-dimensional for indoors. Our method doesn't assume anything about the illumination and surfaces reflectance spectra, and also uses a much simpler geometric computation, which does not require iteratively solving non-linear equations.

Funt and Drew \cite{DBLP:journals/pami/FuntD93} described a simple approach to extract the pure interreflection color. Their model assumes that the colors reside on a plane in the RGB space, so all that is needed in order to recover the interreflection color is to intersect two planes. They use this idea to decompose the image into direct and indirect illumination components. Their work does not suggest to use the interreflection color to derive the illuminant chromaticity, as we do here. Also in real images, colors do not reside on a plane in the RGB space, because of noise. Our method bypasses this complication by sampling multiple pixels to alleviate noise. Another restriction in this paper is that it requires mutual interreflections between pairs of surfaces. Our method can work with two or more unrelated interreflections.

\section{Illuminant Chromaticity Estimation}

In this section we describe our method for extracting illuminant chromaticity from interreflections. After describing the image formation model, we first show how to obtain an estimate of the illuminant chromaticity from a pure interreflection. Next, we consider a more realistic model, where the light reflected off a surface is due to a mixture of direct illumination by the illuminant together with an indirect component (a diffuse interreflection from a nearby surface). We show that two or more such interreflections enable us to estimate the illuminant chromaticity by intersecting lines in a special 2D chromaticity plane.

\subsection{Image formation model}

\begin{figure}
\centering
\includegraphics[width=\columnwidth]{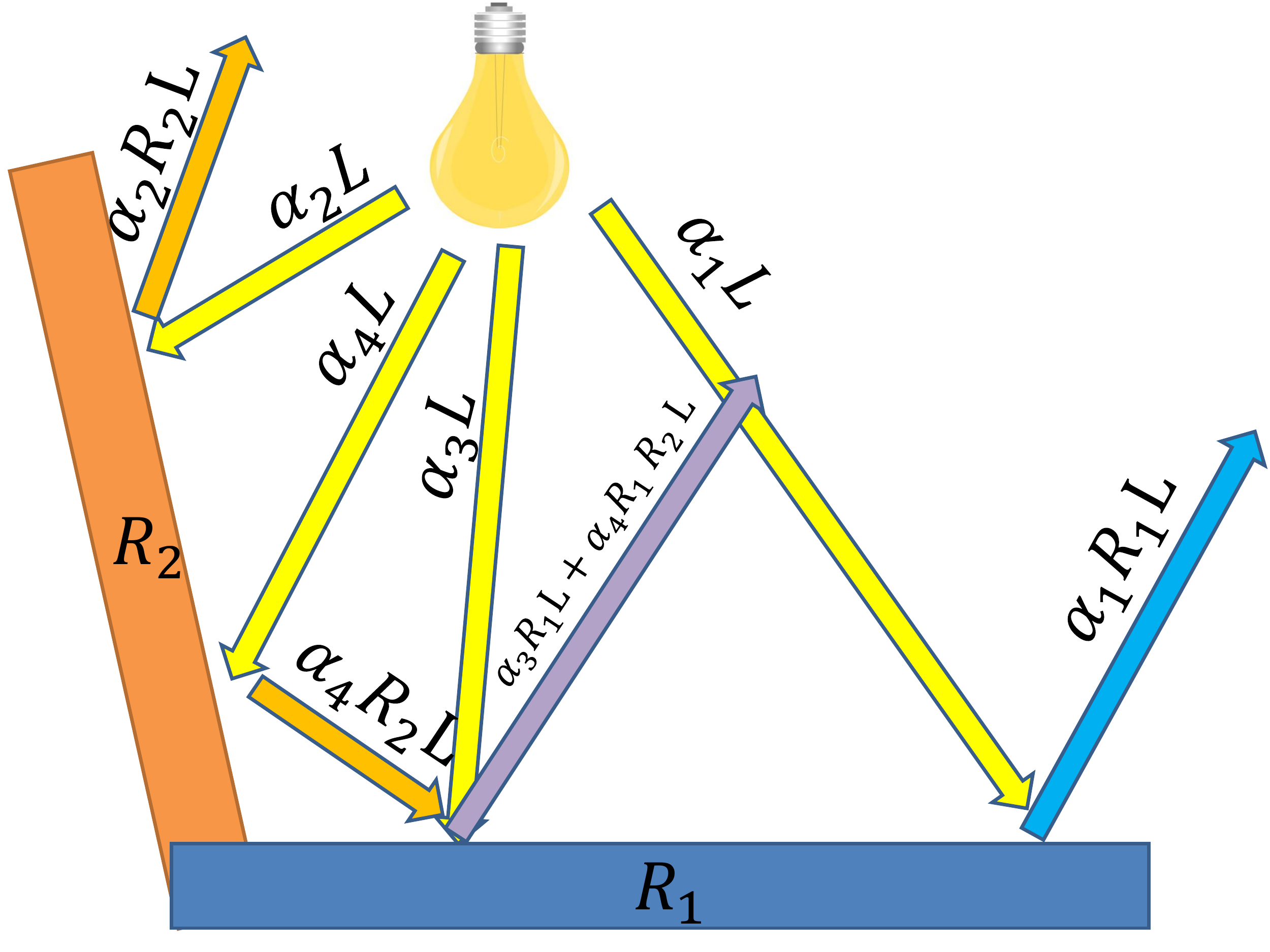}
\caption{\label{image_formation} Interreflection between a pair of adjacent surfaces. Light reflected from the distant parts of the two surfaces is mainly due to direct illumination. Light reflected from the portion of $R_{1}$ closer to $R_{2}$ is due to both direct and interreflected light.}
\end{figure}


Let $L$ be the illuminant color, expressed as a 3D vector in the RGB color space. Since, for our purposes, we are interested only in the chromaticity of the illuminant rather than its intensity, we can assume that the illuminant color $L$ is a unit vector.

Let $R_1$ denote the diffuse reflectance color of a Lambertian surface in the scene, also expressed as a 3D RGB vector. When the above surface is directly illuminated by the illuminant, the reflected color may be expressed as $\alpha_1 R_1 L$, where $R_1 L$ is a pointwise product, and $\alpha_1$ is a scalar that bundles together all the factors that influence light intensity such as shadowing, surface orientation, and illuminant intensity (see Figure \ref{image_formation}). Note that here we effectively assume that the camera RGB sensor responses are Dirac delta functions. We will show later that this assumption does not introduce a significant error, for most real world illuminants and surface reflectances.

Given two adjacent diffuse surfaces, whose reflectances are given by $R_1$ and $R_2$, the light leaving the first surface due to interreflection from the second one may be expressed as $\alpha R_{1}R_{2}L$, for some scalar $\alpha$. In this work, we assume that the interreflected component is significant mainly in the vicinity of the contact between the two surfaces, and can be considered negligible farther away. This is illustrated in Figure \ref{image_formation}.

\subsection{Illuminant estimation from pure interreflection}
\label{sec:pure}

Assume that for the above two surfaces we are able to measure the directly reflected light from each of them, $\alpha_{1} R_{1} L$ and $\alpha_{2} R_{2} L$, as well as the interreflected light leaving one of them, $\alpha_{3} R_{1} R_{2}L$. From these three measurements, we are able to recover the illuminant chromaticity by computing the ratio
\begin{equation} \label{pure_eq}
\frac{\alpha_{1} R_{1} L \cdot \alpha_{2} R_{2}L}{\alpha_{3} R_{1} R_{2} L} = 
\frac{\alpha_{1} \alpha_{2} L}{\alpha_{3}} = \beta L,
\end{equation}
and normalizing the resulting vector $\beta L$. This is merely an approximation, because equation \eqref{pure_eq} is expressed in a 3D color space, rather than using the full illuminant and the surface reflectance spectra, followed by a projection to the RGB color space of the camera. However, the simulations we performed on a data set of real world illuminant and surface reflectance spectra shows this approximation is quite precise, in most cases (see Section \ref{simulated-results}). 

\subsection{Illuminant estimation from a real image}
\label{sec:real}

In real images, we are not able to measure the interreflected component directly, since the surfaces in the scene reflect a combination of direct and indirect illumination. The relative magnitudes of the two components in this combination are not known.
Thus, in order to use equation~\eqref{pure_eq}, it would be necessary to decompose the color of a surface into these two components. Such a decomposition is a challenging problem in itself, and there are some attempts to solve it, for example using special illumination and multiple images, see \eg,
\cite{DBLP:journals/tog/NayarKGR06, DBLP:journals/pami/OTooleMK16, DBLP:journals/tog/OTooleRK12, DBLP:journals/tog/VeltenWJMBJLBGR13}. In this work, instead of trying to decompose the image, we leverage more than one interreflection in order to overcome this ambiguity.



\subsubsection{Color lines}

Assume we have captured two adjacent surfaces under the same illuminant $L$: $\alpha_{1} R_{1} L$ and $\alpha_{2} R_{2} L$ and a portion of the first surface colored by an interreflection, in addition to the direct illumination $L$: $\alpha_{3} R_{1} L + \alpha_{4} R_{1} R_{2} L$, as illustrated in Figure \ref{image_formation}. From these observed quantities, we can compute the following ratio:
\begin{equation} \label{colorBleed}
C = \frac{\alpha_{3} R_{1} L + \alpha_{4} R_{1} R_{2} L}{\alpha_{1} R_{1} L \cdot \alpha_{2} R_{2}L} = 
\frac{\alpha_3}{\alpha_1 \alpha_2} \frac{1}{R_{2}L} + \frac{\alpha_4}{\alpha_1 \alpha_2} \frac{1}{L}
\end{equation}
Note that $C$ is a linear combination of two 3D vectors, $\frac{1}{R_{2}L}$ and $\frac{1}{L}$, whose coefficients are not known to us. The projections of all the possible linear combinations of these two vectors onto a plane in the 3D color space, say, the plane $r+g+b=1$, yields a straight line that contains the projections of $\frac{1}{R_{2}L}$ and $\frac{1}{L}$. We omit the scalars here, since the projection is invariant to multiplication by a scalar.



To conclude, although we are not able to measure the pure interrecflection, we are able to measure $\frac{1}{R_2 L}$ and $C$, and project them onto the plane $r+g+b=1$, thereby obtaining a line that is supposed to pass through the projection of $\frac{1}{L}$ on the same plane.

\begin{figure}
\centering
\includegraphics[width=\columnwidth]{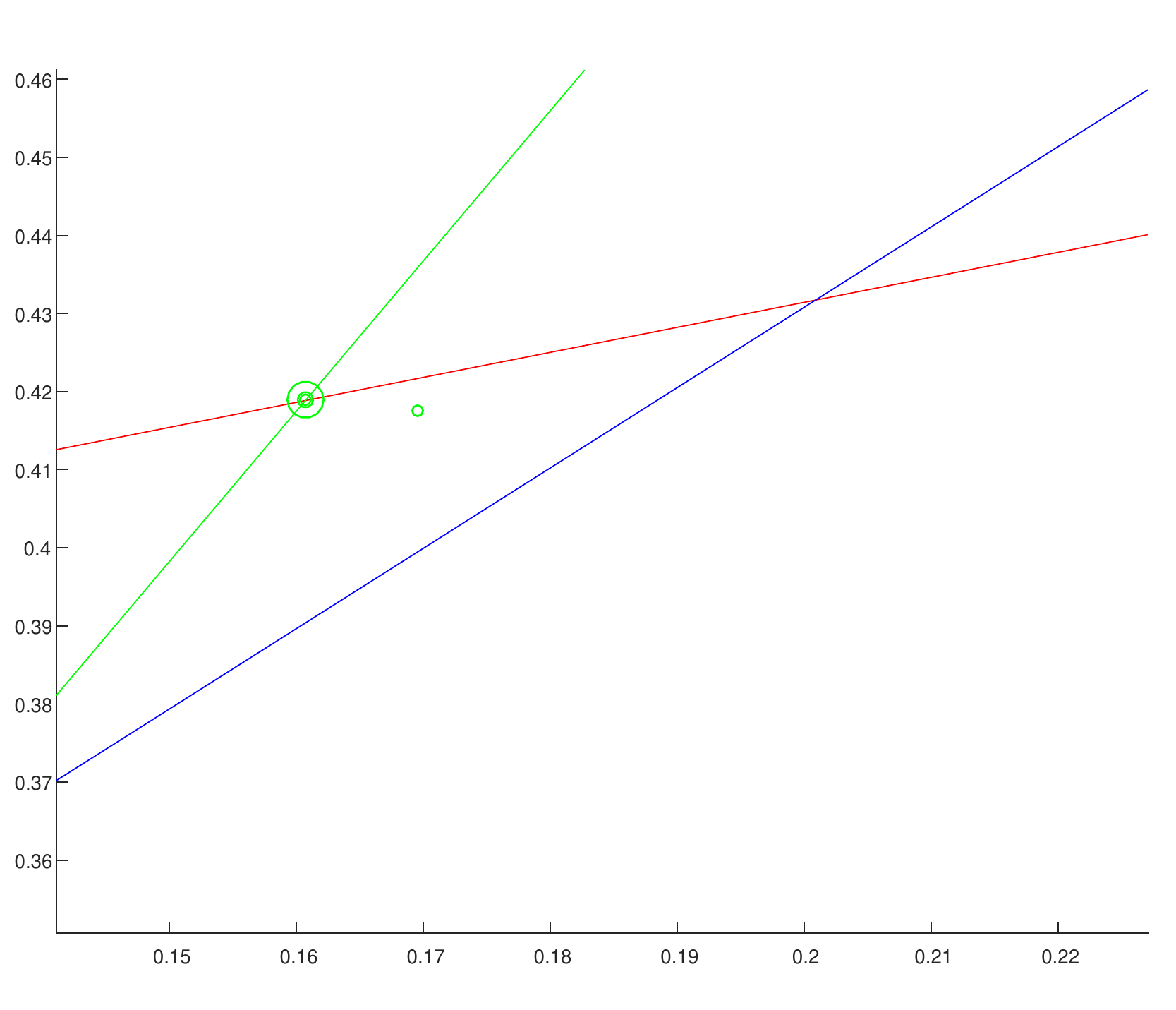}
\caption{\label{little_angle} All three lines are close to $\frac{1}{L}$ (indicated by a green dot). The geometric median (indicated by a double green circle) falls on the intersection with the larger angle, which minimizes the sum of distances to all three lines, rather than intersections with the smaller angles, yielding a better estimate.}
\end{figure}


Next, suppose that we are able to identify at least two areas containing an interreflected component, thereby obtaining two or more color lines, as described above. These lines should all intersect at the projection of $\frac{1}{L}$. Thus, computing their intersection and inverting the resulting coordinates should yield the illuminant chromaticity $L$.
In real images, however, the color lines do not all intersect at the same point.
There are multiple reasons for this, including noise, indirect light coming from other surfaces in the scene, and the use of 3D color vectors instead of continuous spectra in equation \eqref{colorBleed}.
The estimation error could be further magnified if angle between the color lines is small.
We address this issue by using a robust method, inspired by the geometric median, in order to estimate the intersection point.

\subsubsection{Geometric median}

Geometric median is the generalization of median to $n$ dimensions ($n=2$, in our case). Instead of real numbers, we have points $p_{i} \in \mathbb{R}^n$, and the geometric median is defined as the point $q \in \mathbb{R}^n$ which minimizes the sum of distances to all the input points: $q = \underset{x \in \mathbb{R}^n}{\operatorname{arg\,min}} \sum_{i=1}^m \left \| p_i-x \right \|_2$. In the one dimensional case $n=1$, this is just the ordinary median.

It is well known that a median is more robust to outliers than an average. In our case, outliers may arise in regions which contain violations of our assumption that we can reliably measure the quantities in equation~\eqref{colorBleed}, or from a bad pair of surface reflectance spectra (as discussed later, in Section \ref{simulated-results}). In either case, an outlier would correspond to a color line that does not pass near the projection of $\frac{1}{L}$.
Thus we adapt the notion of the geometric median to our setting. Specifically, since we are dealing with straight lines, rather than points, we search for the point $q$ which minimizes the sum of distances to $m$ lines $l_1,\ldots,l_m$, rather than to points:
\begin{equation}
q = \underset{x \in \mathbb{R}^2}{\operatorname{arg\,min}} \sum_{i=1}^m  d(l_i, x).
\end{equation}

Since the geometric median tries to minimize the sum of distances to all the lines, it will inevitably fall on one of the line intersections, except symmetric cases. This approach handles well the case of lines with a small angle, and favors lines with a larger angle, as demonstrated in Figure \ref{little_angle}. Using a least squares formulation, where we seek to minimize the sum of the squares of the distances instead, is more sensitive to outliers, and therefore geometric median overall achieves better results, as can be seen in Table \ref{comparison-table}.


\section{Experiments}

\subsection{Simulations}
\label{simulated-results}

Estimating the illuminant chromaticity from a pure interreflection, using equation \eqref{pure_eq} is an approximation, since it is done in a 3D RGB space, rather than using the full illuminant and reflectance spectra. In order to test the accuracy of this approximation we carried out a simulation. We used the data set presented in \cite{COL:COL10049}, which contains the spectra of 102 illuminants, the reflectance spectra of 1995 materials and the sensor response functions of the camera used in that work. For each illumination spectrum in the dataset, we randomly choose two reflectance spectra (not necessarily different from each other), compute the point-wise multiplication of them with the illuminant spectrum, and project to RGB using the camera response functions. Thus, we obtain a large collection of triplets of RGB vectors $R_1L$, $R_2L$, and $R_1R_2L$. For each triplet, we use equation \eqref{pure_eq} to estimate $L$ and record the angular error \cite{gijsenij2009perceptual} with respect to the ground truth.

\begin{figure}
\centering
\includegraphics[width=\columnwidth]{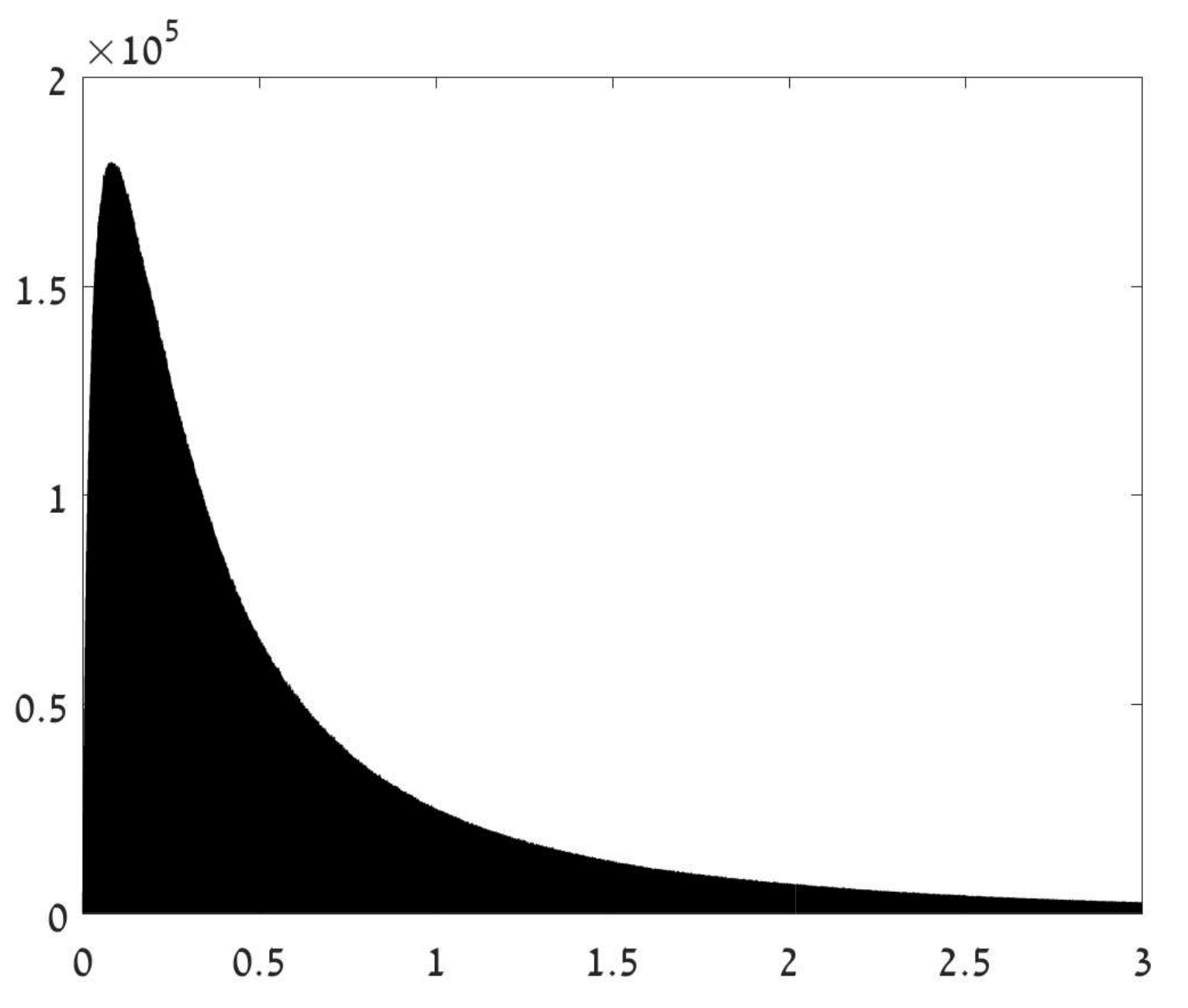}
\caption{\label{pure_cb} Distribution of the angular error of the pure interreflection method, simulated over million random pairs of reflectances per illuminant. The mean error is \degree{0.69} and the median error is \degree{0.37}.
}
\end{figure}

In total, we have sampled one million of random reflectance spectra pairs for each of the 102 illuminants in the dataset. The mean angular error was \degree{0.69} and the median error was \degree{0.37}. The complete angular error distribution is plotted in Figure \ref{pure_cb}, and numerical results are detailed in Table \ref{comparison-table}.


Another simulation was carried out to test the accuracy of the more practical, color-line based approach described in Section \ref{sec:real}. Data for this simulation was produced similarly to the first simulation described above. Here, we also randomly select $\alpha_3$ and $\alpha_4$ to produce the interreflection: $\alpha_{3} R_{1} L + \alpha_{4} R_{1} R_{2} L$, but in fact these values have no effect on the color lines. Instead of sampling one interreflection each time, in this simulation we sample multiple pairs of reflectances, to form multiple color lines, and then compute the geometric median in order to obtain the illuminant estimation. Here, we have also experimented with least squares as an alternative to the geometric median.

\begin{figure}
\centering
\includegraphics[width=\columnwidth]{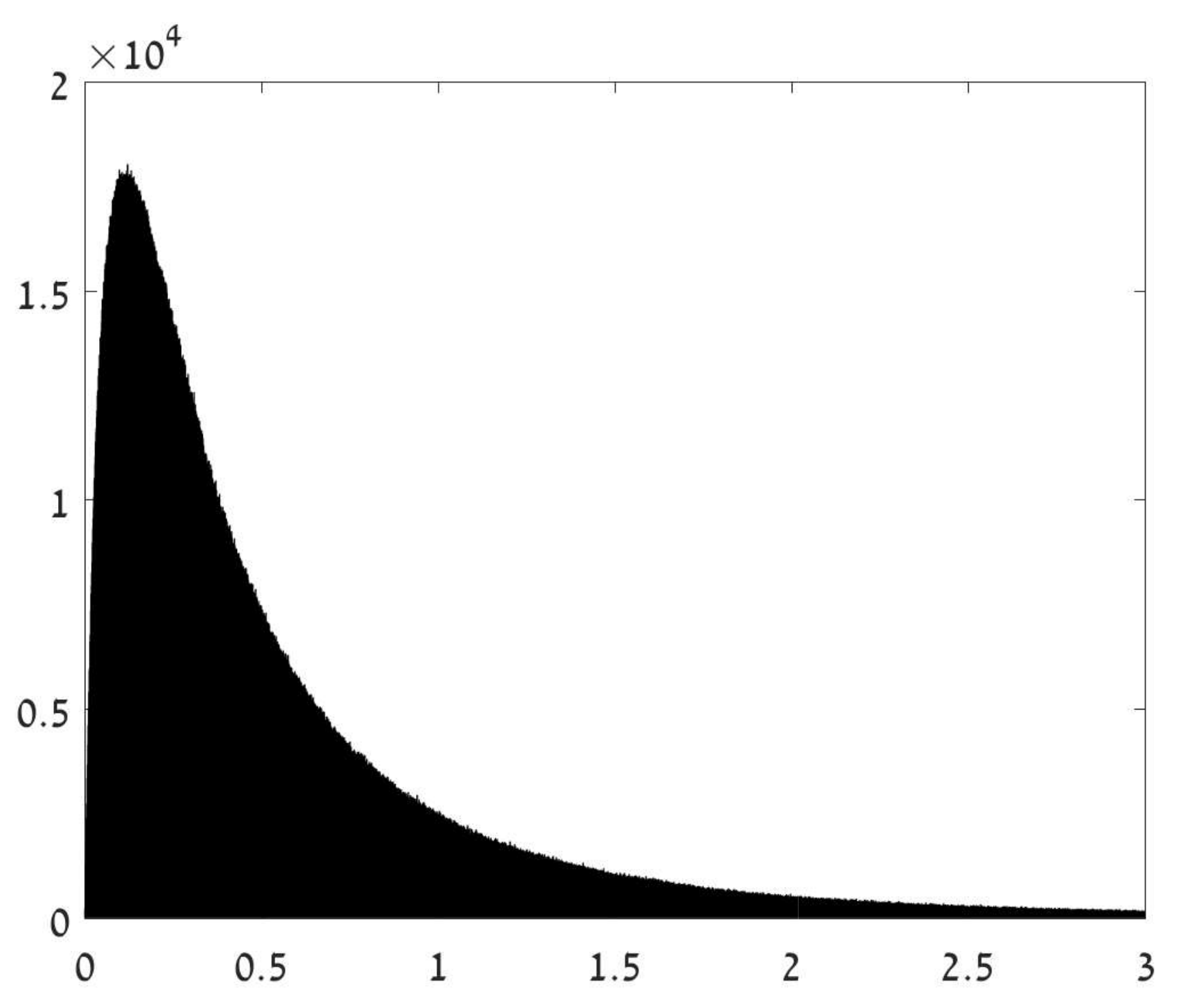}
\caption{\label{5_cb} Distribution of the angular error of the color lines approach, over $10^5$ experiments per illuminant. In each experiment there are five random pairs of surface reflectances, and the geometric median is used to estimate the illuminant chromaticity from five color lines. The median error of \degree{0.36} is similar to the pure interreflection method.}
\end{figure}

Simulations with only two interreflections in each experiment can suffer from large errors, since a large error in any one of the two color lines causes an error of roughly the same magnitude in their intersection point. In addition, a small angle between a pair of lines magnifies the error of each color line. However, increasing the number of color lines drastically reduces the error, and using five color lines produces an average error that is smaller than the one that can be obtained from a single pure interreflection. Specifically, when using five randomly generated interreflections (and using $10^5$ experiments for each of the 102 illuminants), the mean angular error is \degree{0.60} and the median error is \degree{0.36}, as shown in Figure \ref{5_cb} and Table \ref{comparison-table}.

For reference, we also include in Table \ref{comparison-table} the estimation accuracy reported by Shi \etal~\cite{DBLP:conf/eccv/ShiLT16} for their state-of-the-art DS-Net method. Although this is not a direct comparison, since Shi \etal use a dataset of real images, while we use simulations, the significantly better accuracy of our simulations leaves sufficient headroom for our method to achieve better performance on suitable real images as well. This hope for improved performance is further reinforced by the results that we report below, in Section \ref{real-results}, for real images.

\subsection{Real images}
\label{real-results}

\begin{figure}
\centering
\includegraphics[width=\columnwidth]{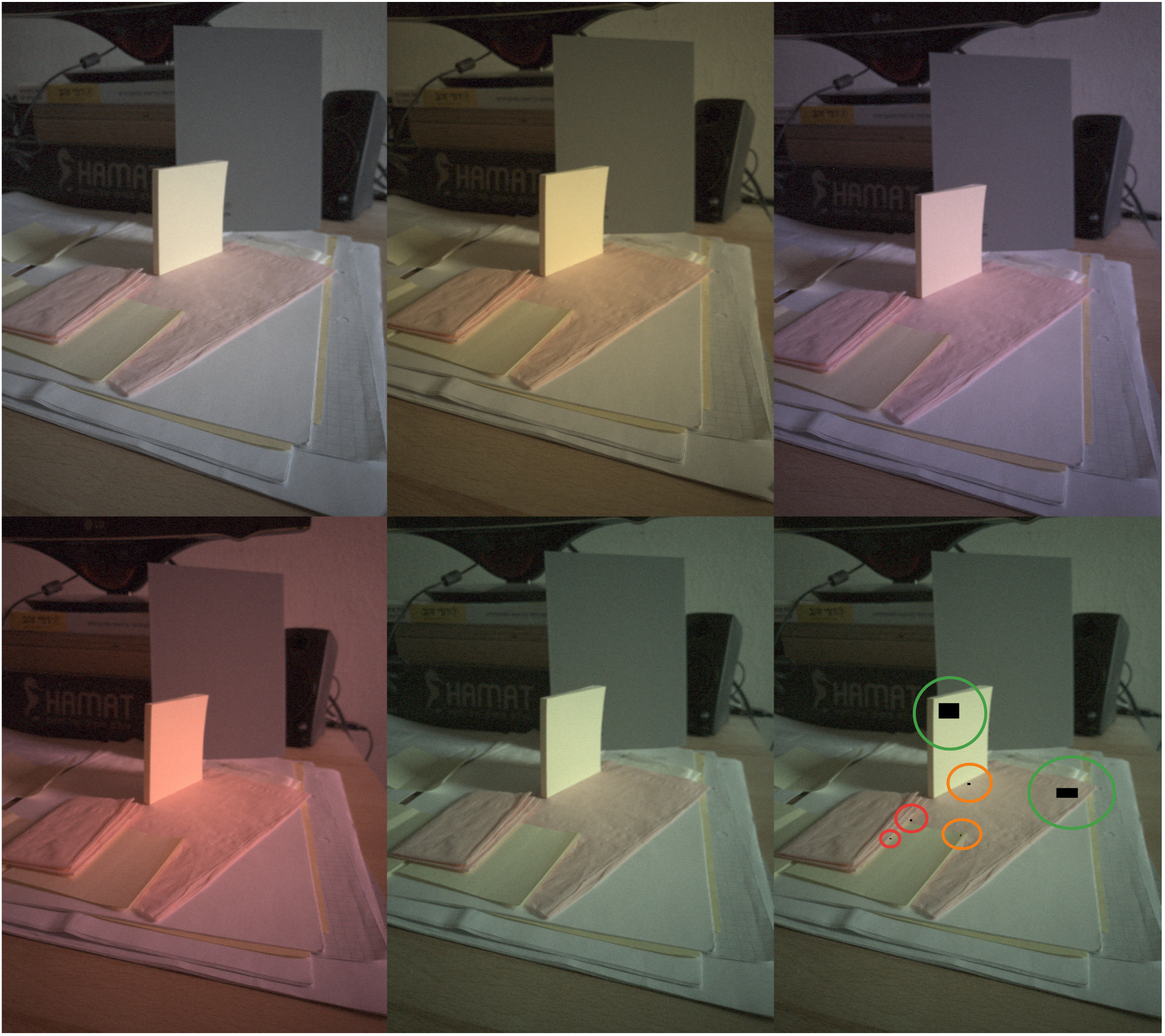}
\caption{\label{images} Real images with color bleeding: The tissue paper and the sticky notes reflect colored light upon each other and upon themselves. The right bottom image is a copy of the middle bottom image, that shows in black the patches that were used to sample the surfaces. The patches used to sample the surface reflectances are indicated by green circles, those which sampled the interreflection caused by color bleed from the pink tissues are indicated by red circles, and those caused by color bleeds from the yellow sticky notes are indicated by the orange circles.
}
\end{figure}

We constructed a real scene in which two surfaces cast a significant interreflection on each other, and photographed this scene under different colored lights using the camera of Nexus 5X. Linear (RAW format) RGB images were recorded. The scene contains a professional gray card from which the ground truth illuminant color may be recovered. The images were captured in a room illuminated only by the daylight coming through a window. We covered the window with colored cellophanes in order to capture the same image with different illuminant colors. The resulting images are shown in Figure~\ref{images}. We manually sampled patches in areas where both surfaces are close to each other, thus making the interreflection significant. We also sampled patches in the far ends of the surfaces, where interreflection are negligible (see the bottom right image in Figure~\ref{images}). In order to eliminate noise we chose patches with high intensity, but we eliminated clipped pixels, and from each patch we computed the median for each RGB channel, and that was the sampled RGB vector provided to our method. The same patch processing was applied when recovering the ground truth illuminant color from the gray card.

\begin{figure}
\centering
\includegraphics[width=\columnwidth]{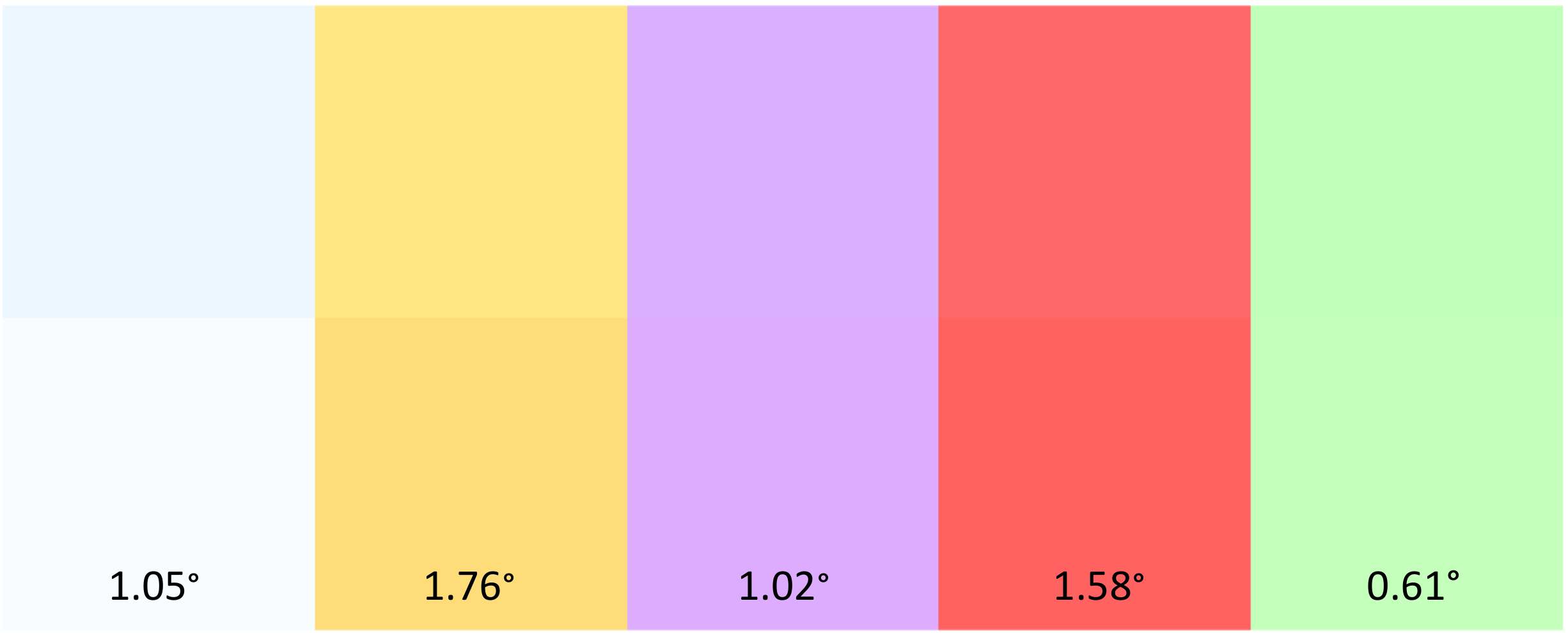}
\caption{\label{estimations} Illuminant color estimation from the real images in Figure \ref{images} using two dominant interreflections. The top row shows the ground truth colors and the bottom row shows our estimates. The angular errors are also reported.
}
\end{figure}

The results are shown in Figure \ref{estimations}. We find that angular error below \degree{1} seem imperceptible, in most cases, as demonstrated by the green illuminant, where the estimation error is \degree{0.61}. For the other illuminants the error is between \degree{1} and \degree{2}.

\begin{figure}
\centering
\includegraphics[width=\columnwidth]{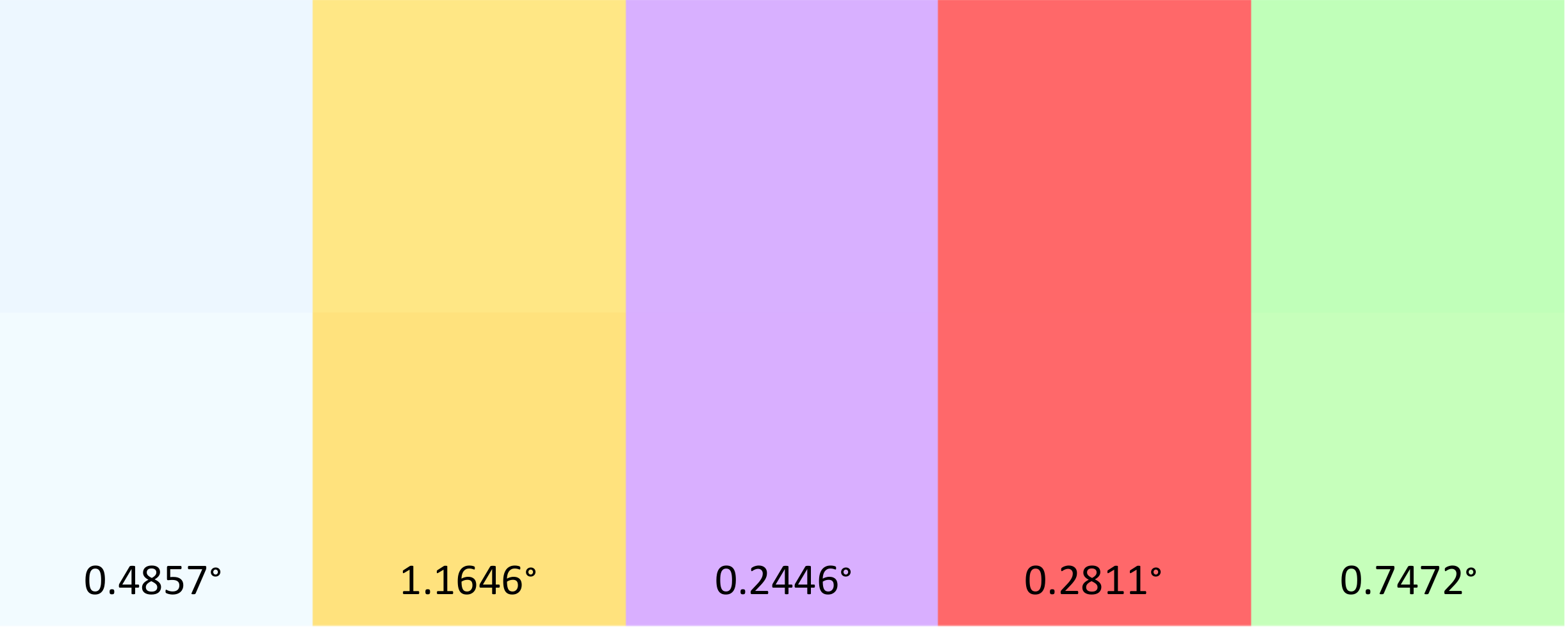}
\caption{\label{estimations_ls} Illuminant color estimation from the real images  in Figure \ref{images} using four interreflections and the least-squares method instead of geometric median. The top row shows the ground truth colors and the bottom row shows our estimates. The angular errors are also reported. 
}
\end{figure}

In order to test the difference between geometric median and least squares on real images, more than two interreflections are needed (in the case of two interreflections both approaches yield the same estimate, which is the intersection of the two lines). There are additional interreflections in our images, between the pink tissue paper and itself, as well as between the yellow sticky note and itself. Using these additional interreflections did not improve the estimation accuracy when using the geometric median, but did reduce the error when using least squares to find the point closest to the color lines. This is shown in Figure \ref{estimations_ls}. Thus, using four interreflections, we are able to achieve an angular error below \degree{1} in four out of the five images.

\begin{figure}
\centering
\includegraphics[width=\columnwidth]{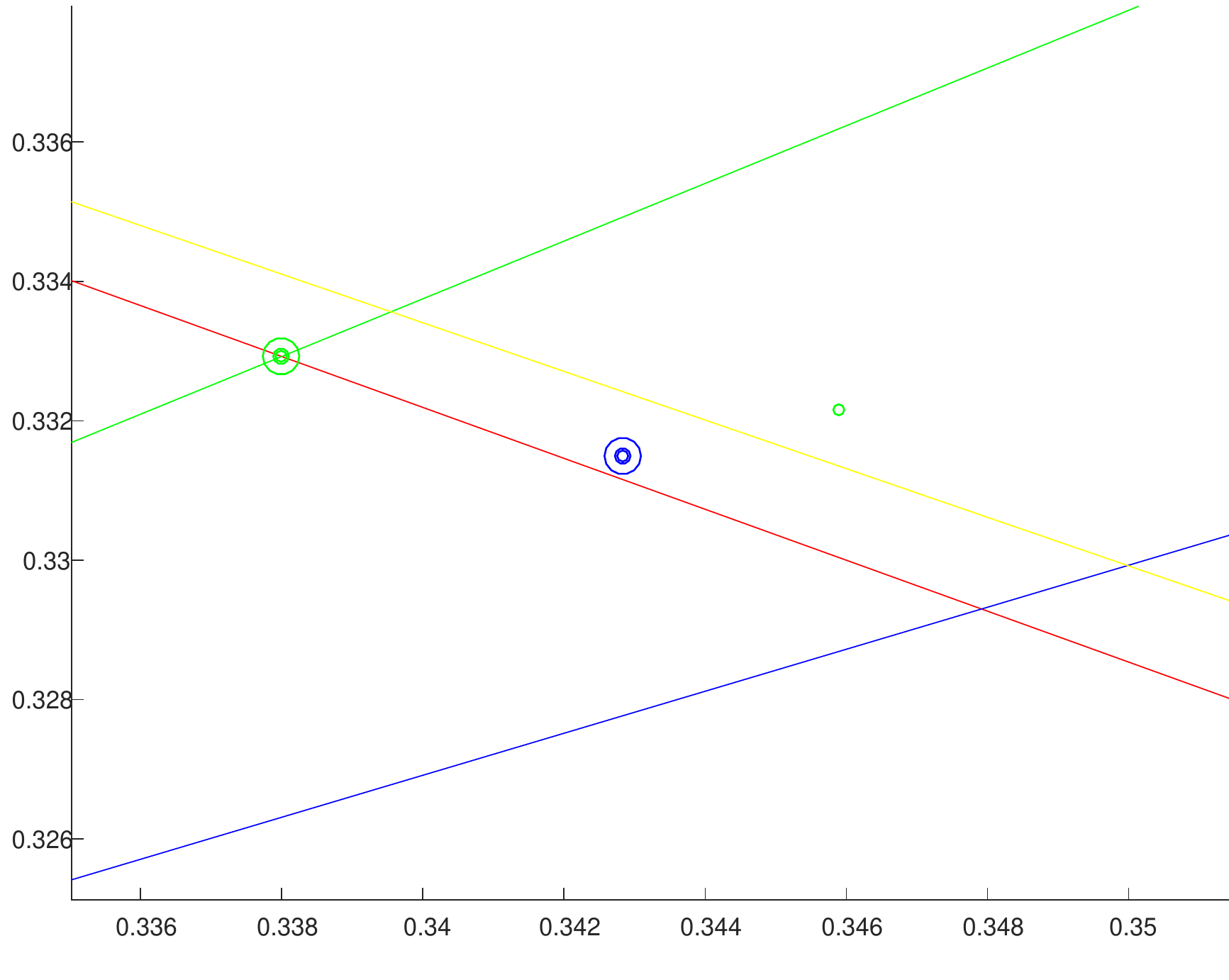}
\caption{\label{symmteric} An example where four lines corresponding to four interreflections, form a roughly symmetric arrangement near the ground truth (green dot). The geometric median falls on one of the intersection points (green circle), which corresponds to a slightly smaller sum of distances than the other intersections. The least squares estimate (blue circle) provides a better estimate in this case.}
\end{figure}

The better performance of the least squares method does not necessarily contradict our simulations, because in these simulations there were also some cases where the least squares method was better than the geometric median. However, its performance was less accurate overall. It seems to be the case that for the real images in Figure~\ref{images}, the ground truth happens to lie between fairly symmetric color lines. While geometric median chooses one of the intersection points, the least squares method yields an average point in the middle which, in this case, happens to be closer to the ground truth. This is depicted in Figure \ref{symmteric}.

\begin{table*}
\centering
\begin{tabular}{|c|c|c|c|c|c|c|c|c|}
\hline
Method & Mean & Median & Trimean & Best-25\% & Worst-25\% & 95th \% & Max & Min \\
\hline
1 pure interreflection & \degree{0.6885} & \degree{0.3713} & \degree{0.4349} & \degree{0.0881} & \degree{1.8445} & \degree{2.4008} & \degree{35.1971} & \degree{0.00002} \\
\hline
2 interreflections & \degree{3.76} & \degree{0.8207} & \degree{1.0283} & \degree{0.1761} & \degree{13.01} & \degree{11.54} & \degree{173.11} & \degree{0.00006} \\
\hline
3 interreflections + GM & \degree{1.26} & \degree{0.5972} & \degree{0.6975} & \degree{0.1435} & \degree{3.57} & \degree{4.1} & \degree{165.97} & \degree{0.00007} \\
\hline
4 interreflections + GM & \degree{0.8129} & \degree{0.4325} & \degree{0.4986} & \degree{0.1127} & \degree{2.19} & \degree{2.71} & \degree{164.15} & \degree{0.000017} \\
\hline
5 interreflections + GM & \degree{0.6088} & \degree{0.3583} & \degree{0.4051} & \degree{0.0978} & \degree{1.56} & \degree{1.96} & \degree{151.04} & \degree{0.00004} \\
\hline
10 interreflections + GM & \degree{0.2940} & \degree{0.2077} & \degree{0.2245} & \degree{0.0638} & \degree{0.6758} & \degree{0.8382} & \degree{8.15} & \degree{0.00012} \\
\hline
3 interreflections + LS & \degree{1.18} & \degree{0.6037} & \degree{0.6890} & \degree{0.1574} & \degree{3.27} & \degree{3.67} & \degree{167.1} & \degree{0.0001} \\
\hline
4 interreflections + LS & \degree{0.8675} & \degree{0.5458} & \degree{0.6062} & \degree{0.1523} & \degree{2.15} & \degree{2.64} & \degree{163.39} & \degree{0.00007} \\
\hline
5 interreflections + LS & \degree{0.7477} & \degree{0.5058} & \degree{0.5540} & \degree{0.1479} & \degree{1.77} & \degree{2.19} & \degree{159.65} & \degree{0.00015} \\
\hline
10 interreflections + LS & \degree{0.5342} & \degree{0.4104} & \degree{0.4367} & \degree{0.1337} & \degree{1.15} & \degree{1.4} & \degree{11.04} & \degree{0.00026} \\
\hline
DS-Net\cite{DBLP:conf/eccv/ShiLT16} & \degree{1.90} & \degree{1.12} & \degree{1.33} & \degree{0.31} & \degree{4.84} & \degree{5.99}  & - & - \\
\hline
\end{tabular}

\vspace{2mm} \caption{\label{comparison-table} Results of our simulations. The top row reports the errors for illuminant estimation from a single pure interreflection, while the following nine rows refer to the color line based approach with various numbers of color lines.
The estimation was done both using the geometric median (GM) and least squares (LS).
For reference, the last row shows the estimation accuracy (on a dataset of real images) reported for the state-of-the-art DS-Net method \cite{DBLP:conf/eccv/ShiLT16}.
}

\end{table*}
\section{Discussion and Future Work}

We assume that the interreflections that we use are between Lambertian surfaces. This is a reasonable assumption in practice, as diffuse surfaces abound in real scenes. Furthermore, when a significant specular reflection is present, it might be easier to recover the illuminant color from specularities. If, on the other hand, the specular reflection is insignificant, then we can neglect it, and treat the surface as Lambertian.

Interreflections exist everywhere around us, however, we usually do not notice them, and they are often too weak to be reliably measured in a photograph. However, capturing interreflections should become more feasible as cameras become more sensitive. Specifically, this should be achievable by the Modulo camera recently introduced by Zhao \etal \cite{DBLP:conf/iccp/ZhaoSFYR15}. Ordinary HDR photos are limited in their capability to capture interreflections, because interreflections are often present in bright regions, and long exposure times saturate the sensor in these regions. However, the Modulo camera is free from this limitation, so one can capture interreflections by using either long exposure or wide aperture.

The weak magnitude of the interreflections actually supports the assumptions that we make in our approach. Their magnitude decreases in inverse proportion to the square of the distance between the two surfaces, making them very local. This makes it possible for us to sample, on the same surface, patches that contain an interreflected component, as well as patches where this component is negligible. It also justifies our decision to adopt the one-bounce model, since the second bounce is even weaker than the first one.

The proposed method should in principle work with other examples of surfaces which emit radiance to their local neighborhood. This could be caused by either interreflection, or by weak local light sources such as LEDs, TVs or smartphone screens. But in the case of light sources, there is a problem of clipping which should be handled first.

While it is unlikely for two random color lines to have a small angle between them, it should be noted that in case of two interreflections reflected from the same surface $R_2$, the lines will almost certainly have a very small angle between them, since according to Equation \eqref{colorBleed}, both lines also go through the point $\frac{1}{R_2L}$. 
This means that the algorithm should avoid the degenerate case of more than one interreflection reflected from the same surface.

\paragraph{Future work} should focus on automating the search for interreflection in the scene. The output of the above search could then be fed as input to the approach described in this paper, yielding a fully automatic accurate method for illuminant chromaticity estimation. The robustness of the geometric median to outliers could be valuable in this scenario, as it will tolerate some amount of wrongly identified interreflections in the image.

{\small
\bibliographystyle{ieee}
\bibliography{egbib}
}

\end{document}